\documentclass{article} 
\usepackage[T1]{fontenc}
\usepackage{iclr2027_conference,times}

\usepackage{amsmath,amsfonts,bm}

\def\eqref#1{equation~\ref{#1}}

\def\1{\bm{1}}

\DeclareMathAlphabet{\mathsfit}{\encodingdefault}{\sfdefault}{m}{sl}
\SetMathAlphabet{\mathsfit}{bold}{\encodingdefault}{\sfdefault}{bx}{n}

\newcommand{\Yset}{\mathcal{Y}}
\newcommand{\D}{\mathcal{D}}
\newcommand{\W}{\mathcal{W}}
\newcommand{\B}{\mathcal{B}}

\usepackage{hyperref}
\hypersetup{hidelinks}
\usepackage{url}

\usepackage{pifont}
\usepackage{xcolor}

\newcommand{\cmark}{\textcolor{green!60!black}{\ding{51}}}
\newcommand{\xmark}{\textcolor{red!75!black}{\ding{55}}}
\usepackage{wrapfig}

\usepackage{booktabs}   
\usepackage{tabularx}   
\usepackage{array}      
\usepackage{multirow}   

\usepackage{graphicx}

\usepackage{tabularx}
\usepackage{array}
\newcolumntype{Y}{>{\centering\arraybackslash}X}

\usepackage{algorithm}
\usepackage{algpseudocode}

\usepackage{needspace}

\newcommand{\para}[1]{\noindent\textbf{#1}}

\title{Label-Centric Knowledge Retrieval for Zero-Shot Text Classification}

\author{
\textbf{Daoping Wu}$^{1,*}$,
\textbf{Dong Wang}$^{1,2,*}$,
\textbf{Siyu Liu}$^{3,*}$,
\textbf{Chen Liang}$^{1,2}$, \\
\textbf{Jian Geng}$^{1,2}$,
\textbf{Ziqi Li}$^{1,2}$,
\textbf{Shuguang Xiong}$^{1}$,
\textbf{Ying Zhang}$^{1}$ \\[1.2em]
$^1$Zhejiang Lab, Hangzhou, China \\
$^2$School of Computer Science and Software Engineering, \\
Harbin Institute of Technology, China \\
$^3$School of Mechanical Engineering, \\
Shanghai Jiao Tong University, China \\[0.8em]
\texttt{daopingwu@outlook.com} \\
\texttt{\{25S103170,23B903050,gengj,24B903030\}@stu.hit.edu.cn} \\
\texttt{liusiyu@sjtu.edu.cn} \\
\texttt{\{xiongshuguang,zhangy\}@zhejianglab.org} \\[0.8em]
{\small $^*$These authors contributed equally to this work.} \\
{\small Dong Wang, Chen Liang, Jian Geng, and Ziqi Li
performed this work during their internships at Zhejiang Lab.}
}

\iclrfinalcopy 
\begin{document}

\maketitle

\begin{abstract}
Flagship large language models (LLMs) offer rich general knowledge and strong reasoning capabilities, but their high inference cost and latency limit their use for large-scale data labeling. We study zero-shot text classification with semantically meaningful labels and task descriptions, without task-specific labeled examples or classifier feedback. We propose \textbf{SLKR} (Semantic Label Knowledge Retrieval), a label-centric retrieval-augmented generation framework that separates offline knowledge construction from lightweight online inference. Offline, a flagship LLM independently generates label definitions, retrieval keywords, and boundary rules for potentially confusable label pairs, forming a reusable label-knowledge store. Because this construction operates over the label space rather than the input corpus, its one-time flagship cost can be amortized across large-scale inference. Online, hybrid semantic and lexical retrieval first identifies the top-$K$ candidate labels and then selects up to $M$ input-relevant boundary rules connecting candidate-label pairs. A lightweight LLM performs the final classification in a single call using the retrieved definitions and rules. Across three public benchmarks and a real-world multilingual book-classification dataset, SLKR achieves the best results on all four datasets with Qwen3.5-4B and on three of four with Ministral-3-3B, while using compact input contexts and no task-specific labeled examples. Boundary ablations further show that retrieving a small set of input-relevant rules preserves performance close to using all boundaries with substantially less context, supporting efficient reuse of flagship-generated label knowledge for lightweight classification.
\end{abstract}

\section{Introduction}
Large language models (LLMs), with their broad knowledge and strong reasoning capabilities, are increasingly used for text annotation and classification. For specialized and broad-coverage data, capable LLMs can outperform non-expert annotators who lack sufficient domain knowledge \citep{gilardi2023chatgpt,bermejo2025llms}. However, stronger models generally come with substantially higher inference costs: closed-source models require expensive API calls and may raise data-privacy concerns, while open-source models demand substantial deployment resources and can be slow at inference. These limitations make repeated use of flagship LLMs impractical for large-scale annotation, particularly in industrial cold-start settings where new classification tasks arise frequently, the scale and diversity of the data make comprehensive labeled coverage difficult, and the conventional pipeline of annotation, human verification, and model fine-tuning introduces substantial iteration overhead.

Importantly, many classification labels already carry meaningful semantics. Labels such as \emph{card payment fee charged} in intent classification, or \emph{Environmental Science and Engineering} and \emph{Civil Engineering} in document classification, are not arbitrary identifiers: their meanings provide a natural interface to the general knowledge encoded in pretrained LLMs. This suggests an opportunity to use a capable LLM to elaborate label semantics once, rather than repeatedly relying on it to classify every input. While prior work has shown that label descriptions can improve classification \citep{kumar2024genz}, we ask a different question: \emph{Can the knowledge of a flagship LLM be organized into reusable label knowledge, enabling cost-efficient classification with a lightweight LLM?}

We introduce SLKR, a label-centric RAG formulation for zero-shot text classification. Conventional corpus-grounded RAG retrieves source passages as evidence \citep{lewis2020retrieval}; our input text instead retrieves candidate classes represented by LLM-augmented label knowledge constructed without task-specific labeled examples. The retrieved content consists of class-level decision guidance rather than task documents or labeled demonstrations, shifting retrieval from corpus evidence to reusable knowledge associated with the label space (Table~\ref{tab:objects}).

\begin{figure}[t]
\centering
\includegraphics[width=\linewidth]{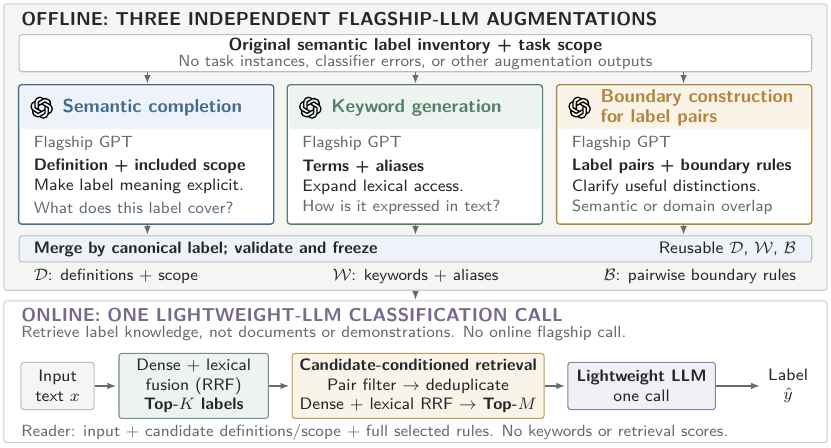}
\caption{\textbf{SLKR overview.} Three independent flagship augmentations receive the same original label inventory and task scope, producing definitions ($\mathcal D$), keywords ($\mathcal W$), and boundary rules ($\mathcal B$). Online, hybrid Top-$K$ label retrieval is followed by candidate-pair filtering, deduplication, and hybrid Top-$M$ rule retrieval. One lightweight LLM reads the input, candidate definitions/scope, and selected rules.}
\label{fig:overview}
\end{figure}

Figure~\ref{fig:overview} gives a high-level overview of SLKR. Offline, a flagship LLM independently augments the label space by enriching label definitions, generating retrieval keywords, and constructing boundary rules for potentially confusing labels. These complementary resources capture label semantics, improve candidate retrieval, and help distinguish ambiguous classes. Online, hybrid retrieval first selects a small set of candidate labels and then retrieves the most relevant boundary rules. A lightweight LLM performs the final classification using only the retrieved label knowledge and the input text. Because expensive flagship-model inference is performed only during label-knowledge construction rather than for each input instance, its cost depends on the label space rather than the input corpus and can be amortized over large-scale inference.

Our evaluation asks whether this division of labor achieves competitive classification quality with shorter contexts and lower recurring costs while accounting for offline construction. We evaluate SLKR on public benchmarks and further validate it in a real-world industrial setting involving large-scale document classification. Our main contributions are summarized as follows.

\Needspace{6\baselineskip}
\para{Contributions.} Our contributions are threefold:
\begin{enumerate}

\item \textbf{A label-centric RAG formulation for zero-shot classification.}
We introduce a retrieval paradigm that retrieves reusable knowledge about candidate labels rather than source documents or labeled task examples. This shifts retrieval from instance-level evidence or demonstrations to class-level decision knowledge, enabling classification without task-specific labeled data.

\item \textbf{Retrieval of explicit decision boundaries between confusable labels.}
We use a capable flagship LLM to identify potentially confusing classes and construct boundary rules that capture distinctions beyond label-name similarity, including semantic, domain, functional, and application-level overlap. At inference time, SLKR retrieves only the boundary knowledge relevant to the input and candidate labels, providing targeted guidance for fine-grained classification.

\item \textbf{Scalable lightweight inference through reusable flagship knowledge.}
We externalize expensive label understanding to offline knowledge construction and reuse it across inputs, leaving online inference to retrieval and a lightweight LLM. This design substantially reduces recurring inference cost and context requirements while maintaining competitive classification quality, making SLKR suitable for large-scale deployment.

\end{enumerate}

\section{Related Work}
\para{Label semantics for classification.}
Prior work has exploited label names and descriptions for text classification. LOTClass and X-Class derive supervision from label names together with unlabeled task corpora \citep{meng2020text,wang2021xclass}, while label-description methods enrich class semantics for zero-shot classification \citep{gao2023benefits,kumar2024genz}. Among these, we compare directly with Gen-Z \citep{kumar2024genz}, which is compatible with our instance-free setting. SLKR further organizes definitions, retrieval keywords, and pairwise boundaries into a reusable label-side knowledge base.

\para{Retrieval-augmented classification.}
Conventional RAG retrieves external documents \citep{lewis2020retrieval}, while retrieved in-context learning selects labeled demonstrations relevant to an input \citep{liu2022what,rubin2022learning}. We include retrieved ICL as a supervised retrieval baseline. Other classification-oriented retrieval methods, such as GORAG \citep{wang2026gorag}, operate in dynamic few-shot settings with labeled data and therefore assume additional task-instance supervision absent from SLKR.

\para{Boundary-aware classification.}
Prior work has used pairwise reasoning to distinguish ambiguous labels \citep{lu2024mitigating}. Confusion-aware retrieval further identifies difficult label pairs from classifier errors and generates targeted decision rules \citep{gupta2026confusion}. These approaches motivate explicit modeling of label boundaries but rely on task instances or classifier feedback. In contrast, SLKR constructs boundary knowledge from the label inventory and task description alone and retrieves input-relevant rules only among the candidate labels.

We next describe how SLKR constructs and retrieves this label-side knowledge for lightweight classification.
\section{Label-Centric Retrieval-Augmented Classification}

We first define the classification setting and the label-side retrieval object. We then describe SLKR in two phases: offline, a flagship LLM independently constructs semantic definitions, retrieval keywords, and boundary rules; online, hybrid label retrieval selects candidate classes, followed by candidate-conditioned boundary retrieval to identify the most relevant decision rules. A lightweight LLM finally predicts from the retrieved label knowledge. Algorithm~\ref{alg:slkr} summarizes the complete pipeline. We conclude with an analysis of how the one-time flagship construction cost is amortized over large-scale inference.

\subsection{Setting and Retrieval Object}

Let $x$ denote an input text and $\Yset=\{y_1,\ldots,y_N\}$ a closed-set inventory of semantically meaningful labels. The task description and any available taxonomy metadata are denoted by $t$. Given $(x,\Yset,t)$, our goal is to predict a primary label $\hat{y}\in\Yset$. Task-specific labeled instances may be used for evaluation and competing few-shot and retrieved in-context learning (ICL) baselines, but are strictly isolated from SLKR's knowledge construction and classifier adaptation.

SLKR uses a frozen flagship LLM $F$ for offline knowledge construction and a frozen lightweight (smaller or lower-cost) LLM $g$ for online classification. For each label $y_i$, $F$ constructs a label record
\[
k_i=(y_i,d_i,w_i,\mathcal{B}_i),
\]
where $d_i$ is an augmented semantic definition, $w_i$ is a set of retrieval keywords, and $\mathcal{B}_i$ contains boundary rules associated with $y_i$. Specifically, a boundary rule $b_{ij}$ describes the decision boundary between a potentially confusable pair $(y_i,y_j)$. The resulting knowledge base
\[
\mathcal{K}=\{k_i\}_{i=1}^{N}
\]
therefore indexes reusable label-side knowledge rather than instances from the corpus being classified.

\begin{table}[t]
\caption{Comparison of prompting and retrieval-based paradigms for LLM classification. 
``No External Data'' denotes no reliance on labeled examples or an external corpus.}
\label{tab:objects}
\centering
\footnotesize
\setlength{\tabcolsep}{3pt}
\renewcommand{\arraystretch}{0.95}

\begin{tabularx}{\linewidth}{
@{}
p{.12\linewidth}
p{.15\linewidth}
>{\centering\arraybackslash}p{.12\linewidth}
>{\centering\arraybackslash}p{.12\linewidth}
>{\centering\arraybackslash}p{.11\linewidth}
X
@{}
}
\toprule
\textbf{Family}
& \textbf{Design}
& \shortstack{\textbf{No External}\\\textbf{Data}}
& \shortstack{\textbf{Label}\\\textbf{Knowledge}}
& \shortstack{\textbf{Boundary}\\\textbf{Aware}}
& \textbf{Provided / Retrieved Content} \\
\midrule

\multirow{2}{*}{Prompting}
& Zero-shot
& \cmark
& \xmark
& \xmark
& Label names and task description \\

& Few-shot
& \xmark
& \xmark
& \xmark
& Text--label demonstrations \\

\midrule

\multirow{3}{*}{\shortstack[l]{Retrieval-\\based}}
& Retrieved ICL
& \xmark
& \xmark
& \xmark
& Retrieved text--label demonstrations \\

& Document RAG
& \xmark
& \xmark
& \xmark
& Retrieved document passages \\

& \textbf{SLKR (ours)}
& \cmark
& \cmark
& \cmark
& Label knowledge and boundary rules \\

\bottomrule
\end{tabularx}
\end{table}

\subsection{Three Independent Flagship Augmentations}

Given the original label inventory $\Yset$ and task description $t$, the flagship LLM $F$ independently constructs three complementary types of label knowledge:
\begin{equation}
\D=F_{\mathrm{sem}}(\Yset,t),\qquad
\W=F_{\mathrm{lex}}(\Yset,t),\qquad
\B=F_{\mathrm{bnd}}(\Yset,t),
\label{eq:augment}
\end{equation}
where the subscripts denote different prompting roles of the same flagship model. Each augmentation is generated directly from $(\Yset,t)$ without using the outputs of the other branches or task-specific labeled examples.

\para{Semantic definitions.}
$F_{\mathrm{sem}}$ expands each label $y_i$ into a concise definition $d_i$ describing its semantic scope and representative topics. For example, for a WoS label such as \textit{Computer Science}, the generated definition may describe the study of computation, algorithms, software, computer systems, artificial intelligence, and related computational methods. This provides richer semantic information than the label name alone.

\para{Retrieval keywords.}
$F_{\mathrm{lex}}$ generates a set of label-specific terms $w_i$ to improve lexical retrieval. For \textit{Computer Science}, these may include \textit{algorithm}, \textit{software}, \textit{machine learning}, \textit{database}, and \textit{computer vision}. The keywords are used only for candidate retrieval and are omitted from the final classification prompt to reduce token usage and avoid redundant context.

\para{Boundary rules.}
Definitions and keywords describe individual labels but do not explicitly resolve ambiguity between related classes. $F_{\mathrm{bnd}}$ therefore identifies potentially confusable label pairs and generates natural-language rules $b_{ij}$ describing how they should be distinguished. For example, a boundary between \textit{Computer Science} and \textit{Electrical Engineering} may distinguish work centered on algorithms, software, and computation from work centered on circuits, electronic devices, and electrical systems. Such boundaries can capture overlap in semantics, domains, methods, and applications that is not apparent from label names alone.

The three outputs are combined into the label-side knowledge base $\mathcal{K}$. Once constructed, $\mathcal{K}$ is frozen and reused across all input instances, allowing the cost of flagship-model augmentation to be amortized over subsequent classification.

\subsection{Hybrid Label Retrieval}

We next describe how the augmented label knowledge is used to retrieve candidate labels. SLKR employs hybrid retrieval that combines semantic and lexical signals. For semantic retrieval, a lightweight frozen encoder $h$ maps the input $x$ and each augmented label definition $d_i$ into the same embedding space, and labels are ranked by cosine similarity. Boundary rules are excluded from label embeddings because they describe contrasts between multiple classes and often contain negative or exclusionary semantics, making them unsuitable as positive semantic representations of individual labels.

In parallel, lexical retrieval matches the input against label names and the LLM-generated keywords $w_i$. These keywords provide direct access to terminology and alternative expressions that may not be captured reliably by semantic similarity alone. They are used only for retrieval and are not included in the final classification prompt, reducing redundant context and token usage.

We combine the semantic and lexical rankings using reciprocal-rank fusion (RRF) \citep{cormack2009reciprocal}:
\begin{equation}
q(i\mid x)
=
\frac{1}{c+\rho_{\mathrm{sem}}(i\mid x)}
+
\frac{\lambda}{c+\rho_{\mathrm{lex}}(i\mid x)},
\qquad
\mathcal{C}_K
=
\operatorname{TopK}_{y_i\in\Yset} q(i\mid x),
\label{eq:rrf}
\end{equation}
where $\rho_{\mathrm{sem}}$ and $\rho_{\mathrm{lex}}$ are the ranks from semantic and lexical retrieval, respectively, $c>0$ is the RRF rank constant, and $\lambda$ controls the contribution of lexical retrieval. The resulting $\mathcal{C}_K$ contains the Top-$K$ candidate labels passed to the boundary-retrieval stage.

\begin{figure}[t]
\centering
\includegraphics[width=\linewidth]{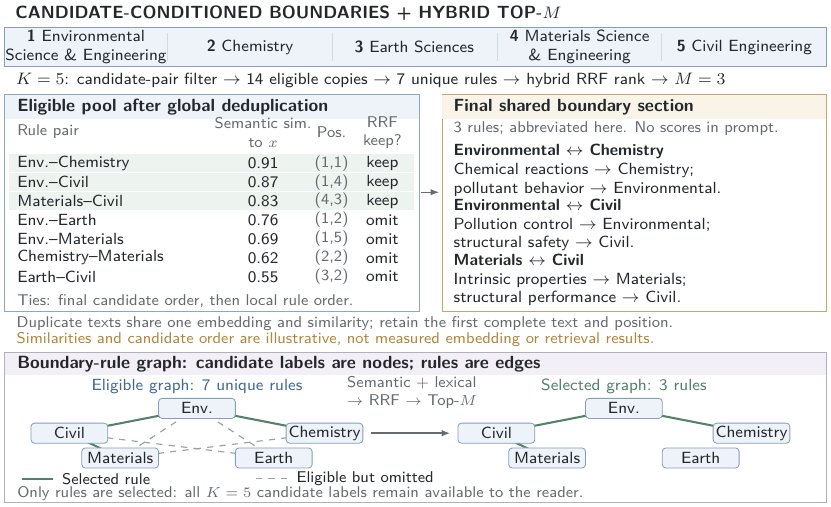}
\caption{\textbf{Candidate-conditioned boundary retrieval} ($K=5$, $M=3$). Fourteen eligible copies yield seven unique rules after deduplication. The table shows illustrative semantic similarities and first-occurrence positions; selection combines semantic and lexical ranks using RRF. The graphs retain all five candidate labels while selecting three rules for shared reader context. Scores and candidate order are schematic, not measured results.}
\label{fig:selection}
\end{figure}

\subsection{Candidate-Conditioned Boundary Retrieval}

The retrieved Top-$K$ labels may be connected by multiple boundary rules, but many of these distinctions may be irrelevant to a particular input. Including all available rules can introduce distracting information and dilute the boundary evidence most relevant to the current classification decision. SLKR therefore performs a second retrieval stage to select only the boundary rules most relevant to the current input.

Given the candidate set $\mathcal{C}_K$, we first restrict the rule pool to boundaries whose two endpoints are both candidate labels:
\begin{equation}
\mathcal{B}(\mathcal{C}_K)
=
\{b_{ij}\mid y_i,y_j\in\mathcal{C}_K,\; i\neq j\}.
\label{eq:eligible}
\end{equation}
Rules involving labels outside $\mathcal{C}_K$ are excluded because they do not help distinguish among the retrieved candidates.

We then apply hybrid retrieval over the eligible boundary rules. Semantic retrieval measures the similarity between the input $x$ and each complete boundary rule using the same lightweight encoder $h$, while lexical retrieval captures direct terminology overlap between the input and the rules. As in label retrieval, the two rankings are combined using RRF, and the Top-$M$ boundary rules are retained:
\begin{equation}
\mathcal{B}_{K,M}(x)
=
\operatorname{TopM}_{b\in\mathcal{B}(\mathcal{C}_K)}
q_{\mathrm{bnd}}(b\mid x),
\label{eq:topm}
\end{equation}
where $q_{\mathrm{bnd}}$ denotes the fused boundary-retrieval score. This input-conditioned selection allows different rules between the same candidate labels to be emphasized for different inputs, reducing irrelevant distinctions while keeping the final context compact.

Figure~\ref{fig:selection} illustrates this process. Label retrieval first defines the candidate subspace; only boundaries connecting candidate labels are then considered, and the most input-relevant rules are selected for final classification.

\begin{algorithm}[t]
\caption{SLKR: Offline augmentation and two-stage retrieval}
\label{alg:slkr}
\small
\begin{algorithmic}[1]
\Require Label inventory $\mathcal{Y}$, task description $t$, flagship LLM $F$, encoder $h$, lightweight LLM $g$, budgets $K,M$
\State \textbf{Offline:} $\mathcal{D} \gets F_{\mathrm{sem}}(\mathcal{Y},t)$; $\mathcal{W} \gets F_{\mathrm{lex}}(\mathcal{Y},t)$; $\mathcal{B} \gets F_{\mathrm{bnd}}(\mathcal{Y},t)$
\State Construct and freeze $\mathcal{K} \gets (\mathcal{D},\mathcal{W},\mathcal{B})$
\State \textbf{Online:} $\mathcal{C}_K \gets \mathrm{HybridRetrieve}(x,\mathcal{D},\mathcal{W},h,K)$
\Comment{Top-$K$ labels}
\State $\mathcal{B}(\mathcal{C}_K) \gets
\{b_{ij} \in \mathcal{B} \mid y_i,y_j \in \mathcal{C}_K,\ i\neq j\}$
\State $\mathcal{B}_{K,M}(x) \gets
\mathrm{HybridRetrieve}(x,\mathcal{B}(\mathcal{C}_K),h,M)$
\Comment{Top-$M$ rules}
\State $\hat{y} \gets
g\!\left(
x,
\{(y_i,d_i) \mid y_i \in \mathcal{C}_K\},
\mathcal{B}_{K,M}(x)
\right)$
\State \Return $\hat{y}$
\end{algorithmic}
\end{algorithm}

\subsection{Lightweight Classification}

After the two retrieval stages, the final classification is performed by the lightweight LLM $g$. The model receives the input text $x$, the names and definitions of the retrieved Top-$K$ candidate labels, and the selected Top-$M$ boundary rules:
\begin{equation}
\hat y
=
g\!\left(
x,
\{(y_i,d_i):y_i\in\mathcal{C}_K\},
\mathcal{B}_{K,M}(x)
\right).
\label{eq:reader}
\end{equation}
The retrieved definitions provide the semantic meaning of the candidate labels, while the boundary rules provide input-relevant guidance for distinguishing potentially confusable candidates. The lightweight LLM then outputs a single predicted label $\hat y\in\mathcal{C}_K$. Retrieval scores and keyword lists are used only during retrieval and are not included in the final prompt, avoiding redundant context and reducing token usage.

\subsection{Amortized Cost Analysis}
\label{sec:cost}

A key motivation of SLKR is that expensive flagship-model computation depends on the label space rather than the number of instances to be classified. Let $N=|\mathcal{Y}|$ denote the number of labels and $T$ the number of input instances. Semantic-definition and keyword generation require $\mathcal{O}(N)$ label-level augmentation, while pairwise boundary construction is upper-bounded by $\mathcal{O}(N^2)$. Thus, the one-time flagship construction cost is $C_{\mathrm{offline}}=\mathcal{O}(N^2)$ in the worst case and, importantly, is independent of $T$.

After construction, each input requires only retrieval and one lightweight-LLM call. Let $c_{\mathrm{online}}$ denote the average online cost per instance. The total and amortized per-instance costs are
\begin{equation}
\begin{aligned}
C_{\mathrm{SLKR}}(T)
&= C_{\mathrm{offline}} + T c_{\mathrm{online}},\\
\bar{C}_{\mathrm{SLKR}}(T)
&= \frac{C_{\mathrm{offline}} + T c_{\mathrm{online}}}{T}
 = \frac{C_{\mathrm{offline}}}{T} + c_{\mathrm{online}}.
\end{aligned}
\label{eq:amortized_cost}
\end{equation}

For a fixed label inventory, the one-time flagship cost is distributed across an increasing number of inputs. Consequently,
\begin{equation}
\lim_{T\rightarrow\infty}
\frac{C_{\mathrm{offline}}}{T}=0,
\qquad
\lim_{T\rightarrow\infty}
\bar{C}_{\mathrm{SLKR}}(T)=c_{\mathrm{online}}.
\label{eq:amortized_limit}
\end{equation}

Thus, at scale, the per-instance contribution of flagship augmentation vanishes, leaving retrieval and lightweight inference as the recurring cost. In contrast, direct flagship classification requires a flagship-model call for every input, so its expensive inference cost grows linearly with $T$.

\section{Experimental Setup}
We evaluate SLKR on classification quality, context efficiency, and amortized inference cost. We additionally study the contribution of boundary rules through ablation.

\subsection{Datasets}

We evaluate SLKR on three public benchmarks and one real-world industrial dataset, summarized in Table~\ref{tab:datasets}. \textbf{WOS-46985} \citep{kowsari2017hdltex,kowsari2018wos} contains scientific abstracts organized in a hierarchical taxonomy; we use the fine-grained Level-2 research areas as classification labels, resulting in 143 classes. \textbf{BANKING77} \citep{casanueva2020efficient} contains 77 fine-grained banking intents with substantial semantic overlap between classes. \textbf{CLINC150} \citep{larson2019evaluation} contains 150 intent classes spanning multiple domains; we use its in-scope examples for classification.

\textbf{Book-38} is a real-world industrial dataset for large-scale book classification. It contains 719 multilingual books categorized into 38 broad disciplines, with 540 training and 179 test instances. Compared with the public benchmarks, Book-38 contains longer and more heterogeneous documents across a broader range of subject areas, providing a complementary evaluation in a practical data-curation setting.

\begin{table}[t]
\caption{Datasets, label counts, evaluation splits, and roles.}
\label{tab:datasets}
\small
\begin{tabularx}{\linewidth}{@{}l r >{\raggedright\arraybackslash}p{.37\linewidth}Y@{}}
\toprule
Dataset & Labels & Split & Role\\
\midrule
WOS-name-143
& 143
& 28,191 train; 9,427 dev; 9,367 test
& Scientific documents; domain overlap\\

BANKING77
& 77
& 10,003 train; 3,080 test
& Fine-grained banking intents\\

CLINC150
& 150
& 15,000 train; 3,000 dev; 4,500 test
& In-scope intent classification\\

Book-38
& 38
& 540 train; 179 test (719 total)
& Multilingual books; real-world industrial classification\\
\bottomrule
\end{tabularx}
\end{table}

\subsection{Baselines and Implementation}

We compare SLKR against representative zero-shot, few-shot, and retrieval-based classification methods. \textbf{Label-only ZS} directly prompts the lightweight LLM with the label inventory. \textbf{Gen-Z} \citep{kumar2024genz} augments each label with a semantic description for zero-shot classification. \textbf{1-shot/label} provides one labeled demonstration per class, while \textbf{Retrieved ICL} retrieves input-relevant labeled demonstrations from the training set. \textbf{SLKR w/o Boundary} uses the same label-retrieval pipeline as SLKR but removes boundary rules ($M=0$), serving as an ablation of boundary retrieval. The full \textbf{SLKR} uses $K=15$ candidate labels and retrieves up to $M=5$ boundary rules.

All retrieval-based methods use Qwen3-Embedding-0.6B \citep{zhang2025qwen3embedding} as the embedding model. We evaluate two lightweight readers from different model families, Qwen3.5-4B \citep{qwen2026qwen35} and Ministral-3-3B \citep{liu2026ministral3}. The offline label knowledge used by SLKR is constructed with GPT-5.6 Sol \citep{openai2026gpt56sol} and frozen across online experiments. Importantly, knowledge construction uses only the label inventory and task description, with no access to training, development, or test instances or classifier feedback. In contrast, few-shot and retrieved ICL baselines use labeled examples from the training split. Each experiment is repeated with three random seeds, and we report the mean performance across runs.

\Needspace{8\baselineskip}
\begin{wraptable}{r}{0.48\linewidth}
\vspace{-8pt}
\centering
\caption{First-stage label retrieval recall (\%).}
\label{tab:retrieval_recall}
\scriptsize
\setlength{\tabcolsep}{2.5pt}
\begin{tabular}{@{}lrrrr@{}}
\toprule
Dataset & R@3 & R@5 & R@10 & R@15 \\
\midrule
BANKING77 & 74.74 & 83.51 & 94.29 & 97.18 \\
CLINC150  & 86.24 & 92.38 & 96.47 & 97.47 \\
WOS-143   & 79.49 & 87.81 & 94.45 & 95.90 \\
Book-38   & 73.74 & 81.38 & 97.39 & 98.32 \\
\bottomrule
\end{tabular}
\vspace{-6pt}
\end{wraptable}

\para{Candidate budget.}
We use $K=15$ for SLKR across all datasets. As shown in Table~\ref{tab:retrieval_recall}, Recall@15 exceeds 95\% on all four datasets, indicating that a relatively small candidate set retains the correct label for the vast majority of inputs. We therefore fix $K=15$ for all subsequent experiments.

\subsection{Boundary Retrieval Ablation}

We evaluate boundary retrieval under four settings while fixing the candidate budget at $K=15$. \textbf{No Boundary} ($M=0$) removes boundary rules entirely. \textbf{All Boundaries} includes all eligible rules connecting the retrieved candidate labels. \textbf{Random $M=5$} randomly selects five eligible boundary rules, providing a rule-count-matched control. \textbf{Hybrid Top-$M=5$}, our default setting, selects the five most input-relevant rules using the hybrid boundary retriever. These settings isolate the effect of boundary knowledge and whether input-conditioned rule selection provides additional benefit over including or randomly selecting boundary rules.

\subsection{Metrics}

We evaluate classification performance using accuracy and macro-F1. For the first-stage label retriever, we report Recall@$K$, which measures whether the gold label is retained among the Top-$K$ candidates. Context efficiency is measured by the average number of input tokens supplied to the lightweight reader, while boundary-retrieval experiments additionally report the average number of tokens contributed by selected boundary rules. For cost evaluation, we separately account for the one-time flagship knowledge-construction cost and the recurring online inference cost, following the amortized analysis in Section~\ref{sec:cost}.

\section{Results}

\para{Main results.}
Table~\ref{tab:main} shows that SLKR achieves the best results on all four datasets with Qwen3.5-4B and on three of four datasets with Ministral-3-3B; retrieved ICL is slightly better on CLINC150 with Ministral. The gains are substantially larger for the weaker reader, suggesting that externalized label knowledge can compensate for limited model capacity. SLKR also compares favorably with few-shot methods despite requiring no task-specific labeled examples, while using only 1.4k input tokens on average compared with 4.7k for 1-shot-per-label prompting. Together, these results support the intended division of labor: expensive label understanding is performed offline, while lightweight models can reuse this knowledge with compact online context.

\para{Boundary retrieval.}
Table~\ref{tab:boundary} shows that explicit boundary knowledge generally improves over definitions alone, but selecting relevant rules is important. Random $M=5$ consistently underperforms Hybrid Top-$M=5$, indicating that the gain does not come simply from adding more rule text. Hybrid retrieval uses only 0.35k boundary tokens, compared with 2.51k when all boundaries are included, while retaining similar performance on the public benchmarks. On the more heterogeneous Book-38 dataset, selective retrieval performs substantially better than including all boundaries, suggesting that filtering irrelevant distinctions becomes particularly useful when class boundaries and document content are more diverse.

\begin{table}[t]
\caption{Classification accuracy / macro-F1 (\%). Bold marks the best results within each model. Input tokens are averaged across the three public datasets.}
\label{tab:main}
\scriptsize
\setlength{\tabcolsep}{2pt}
\renewcommand{\arraystretch}{0.92}
\begin{tabularx}{\linewidth}{@{}lYccccc@{}}
\toprule
Model & Method & WOS-143 & BANKING77 & CLINC150 & Book-38 &
\shortstack{Avg. Input\\Tokens}\\
\midrule

\multirow{6}{*}{Qwen3.5-4B}
& Label only -- ZS
& 57.10 / 56.54 & 68.77 / 66.99 & 80.36 / 78.94 & 64.25 / 57.04 & 0.9k\\
& Gen-Z (Label + definition)
& 57.37 / 57.06 & 72.95 / 71.58 & 85.64 / 84.66 & 61.45 / 60.81 & 2.7k\\
& 1-shot / label
& 58.67 / 57.88 & 74.84 / 73.86 & 89.07 / 88.66 & 68.16 / 65.57 & 4.7k\\
& Retrieved ICL
& 59.67 / 57.92 & 75.29 / 74.40 & 89.24 / 88.92 & 67.60 / 68.08 & 1.2k\\
& SLKR-Desc ($M=0$)
& 59.92 / 57.67 & 77.05 / 75.99 & 89.13 / 88.57 & 66.85 / 66.24 & 1.0k\\
& SLKR ($M=5$)
& \textbf{60.17} / \textbf{58.20}
& \textbf{77.45} / \textbf{76.63}
& \textbf{89.81} / \textbf{89.33}
& \textbf{70.58} / \textbf{70.25}
& 1.4k\\

\midrule

\multirow{6}{*}{Ministral-3-3B}
& Label only -- ZS
& 29.04 / 29.76 & 61.17 / 59.96 & 72.64 / 71.08 & 39.29 / 35.74 & 0.9k\\
& Gen-Z (Label + definition)
& 34.49 / 35.55 & 65.91 / 64.79 & 81.36 / 80.35 & 46.37 / 44.87 & 2.7k\\
& 1-shot / label
& 42.05 / 42.00 & 71.62 / 70.28 & 85.31 / 84.69 & 54.19 / 46.62 & 4.7k\\
& Retrieved ICL
& 46.86 / 45.76 & 70.26 / 68.59
& \textbf{86.07} / \textbf{85.60}
& 54.75 / 47.65 & 1.2k\\
& SLKR-Desc ($M=0$)
& 52.93 / 51.31 & 67.76 / 66.70 & 84.00 / 83.22 & 53.07 / 49.93 & 1.0k\\
& SLKR ($M=5$)
& \textbf{55.83} / \textbf{53.75}
& \textbf{73.44} / \textbf{72.81}
& 85.69 / 85.07
& \textbf{58.66} / \textbf{52.29}
& 1.4k\\

\bottomrule
\end{tabularx}
\end{table}

\begin{table}[t]
\caption{Boundary retrieval ablation at $K=15$. Each cell reports accuracy (top) and macro-F1 (bottom), in \%. Boundary tokens denote the average per-input boundary context.}
\label{tab:boundary}
\scriptsize
\setlength{\tabcolsep}{2pt}
\begin{tabularx}{\linewidth}{@{}Yccccc@{}}
\toprule
Boundary policy & WOS-143 & BANKING77 & CLINC150 & Book-38 & \shortstack{Boundary\\tokens}\\
\midrule
No boundary ($M=0$)
&\shortstack{$59.92 \pm 0.07$\\$57.67 \pm 0.07$}
&\shortstack{$77.05 \pm 0.00$\\$75.99 \pm 0.01$}
&\shortstack{$89.13 \pm 0.03$\\$88.57 \pm 0.03$}
&\shortstack{$66.85 \pm 0.32$\\$66.24 \pm 1.62$}
&0\\

\midrule
All boundaries
&\shortstack{$\mathbf{60.36} \pm 0.08$\\$\mathbf{58.34} \pm 0.04$}
&\shortstack{$\mathbf{77.47} \pm 0.17$\\$\mathbf{76.82} \pm 0.20$}
&\shortstack{$\mathbf{89.95} \pm 0.25$\\$\mathbf{89.44} \pm 0.25$}
&\shortstack{$67.04 \pm 0.00$\\$64.71 \pm 0.18$}
&2.51k\\

\midrule
Random $M=5$
&\shortstack{$58.72 \pm 0.35$\\$56.92 \pm 0.31$}
&\shortstack{$76.69 \pm 0.06$\\$75.78 \pm 0.09$}
&\shortstack{$89.41 \pm 0.18$\\$88.88 \pm 0.17$}
&\shortstack{$66.48 \pm 0.56$\\$62.55 \pm 2.04$}
&0.36k\\

\midrule
Hybrid Top-$M=5$
&\shortstack{$60.17 \pm 0.05$\\$58.20 \pm 0.04$}
&\shortstack{$77.45 \pm 0.02$\\$76.63 \pm 0.02$}
&\shortstack{$89.81 \pm 0.03$\\$89.33 \pm 0.03$}
&\shortstack{$\mathbf{70.58} \pm 0.32$\\$\mathbf{70.25} \pm 0.17$}
&0.35k\\
\bottomrule
\end{tabularx}
\end{table}

\begin{table}[!t]
\caption{Flagship--lightweight cost comparison on BANKING77. Amortized cost is computed over three complete BANKING77 test-set runs (9240 classifications in total). Costs are reported in USD.}
\label{tab:builder}
\footnotesize
\setlength{\tabcolsep}{3pt}
\begin{tabularx}{\linewidth}{@{}l ll YYYY@{}}
\toprule
Method & Builder & Reader &
\shortstack{Accuracy\\ / F1} &
\shortstack{Build\\Cost} &
\shortstack{Online Cost\\ / 1k} &
\shortstack{Amortized\\Cost / 1k}\\
\midrule
Direct Flagship & None & GPT-5.6-sol
& 86.01 / 85.30 & 0 & 4.111 & 4.111\\
\textbf{SLKR} & GPT-5.6-sol & Qwen-3.5-4B
& 77.45 / 76.63 & 3.4 & 0.0596 & 0.4276\\
\bottomrule
\end{tabularx}
\end{table}
\para{Flagship and cost comparison.}
Table~\ref{tab:builder} compares SLKR with direct GPT-5.6 Sol classification on BANKING77. Direct flagship inference achieves higher classification quality (86.01/85.30 accuracy/macro-F1 versus 77.45/76.63), establishing a clear quality--cost trade-off. However, its inference cost is incurred for every input, whereas SLKR pays the flagship cost only once to construct reusable label knowledge and performs online classification with Qwen3.5-4B. The resulting recurring online cost is substantially lower (\$0.0596 versus \$4.111 per 1k inputs). As the workload grows, the fixed construction cost is amortized across inputs, consistent with the analysis in Section~\ref{sec:cost}.

\section{Conclusion}

We presented SLKR, a label-centric retrieval framework for zero-shot text classification that externalizes expensive label understanding into reusable knowledge constructed by a flagship LLM. Rather than retrieving documents or labeled demonstrations, SLKR retrieves candidate-label knowledge and input-relevant boundary rules, enabling lightweight LLMs to classify with compact contexts and without task-specific labeled examples. Across three public benchmarks and a real-world multilingual book-classification dataset, SLKR consistently improves lightweight zero-shot classification and compares favorably with few-shot and retrieved-ICL baselines across two model families. Boundary ablations further show that selecting a small number of relevant rules can retain the benefit of much larger boundary contexts while avoiding irrelevant distinctions. By shifting flagship computation from per-instance inference to reusable offline label knowledge, SLKR provides a practical path toward scalable and cost-efficient classification in large-scale cold-start settings.

\section*{AI Use Statement}

Generative AI tools were used to polish the writing of this paper and to assist with code development and debugging. All AI-assisted content and code were reviewed and verified by the authors. The authors take full responsibility for the final content of this work.

\clearpage

\bibliography{references}
\bibliographystyle{iclr2027_conference}

\appendix
%

\appendix

\section{Additional Method Details}
\label{app:method}

\subsection{Independent Label-Knowledge Construction}
\label{app:construction}

SLKR constructs three complementary types of label-side knowledge with a
flagship LLM: semantic definitions, retrieval keywords, and pairwise boundary
rules. As described in Section~3.2, the three augmentation branches are
independent. Each branch receives the original label inventory and task
description, without using outputs from the other branches or task-specific
labeled examples.

Semantic completion expands each label into a concise definition describing its
semantic scope and representative topics. Keyword generation produces
label-specific terms that are used for lexical candidate retrieval. Boundary
construction identifies potentially confusable label pairs and produces
natural-language rules describing distinctions between them.

After construction, the three resources are combined into the label-side
knowledge base and frozen for online inference. The one-time flagship
construction is therefore performed over the label space rather than separately
for every input instance.

\subsection{Hybrid Candidate-Label Retrieval}
\label{app:label-retrieval}

The first online stage retrieves candidate labels using both semantic and
lexical signals. A frozen lightweight encoder maps the input and augmented label
definitions into the same embedding space, and semantic ranking is based on
cosine similarity. In parallel, lexical retrieval uses label names and
LLM-generated keywords.

The semantic and lexical rankings are combined using reciprocal-rank fusion
(RRF). Boundary rules are excluded from the individual label embeddings because
they describe contrasts between multiple labels rather than positive semantic
representations of a single class. Generated keywords are used for retrieval
but are not included in the final lightweight-LLM classification context.

The main experiments use a candidate budget of $K=15$ for all datasets.
Table~3 in the main paper reports Recall@15 of 97.18\% on BANKING77, 97.47\%
on CLINC150, 95.90\% on WOS-name-143, and 98.31\% on Book-38.

\subsection{Candidate-Conditioned Boundary Retrieval}
\label{app:boundary-retrieval}

After candidate-label retrieval, SLKR restricts the boundary-rule pool to rules
whose two label endpoints both occur in the retrieved Top-$K$ candidate set.
Rules involving labels outside the candidate set are excluded.

The eligible boundary rules are then retrieved using hybrid semantic and lexical
signals. Semantic retrieval measures similarity between the input and each
complete boundary rule using the same lightweight encoder, while lexical
retrieval captures direct terminology overlap. The two rankings are combined
using RRF, and the Top-$M$ boundary rules are retained.

The main configuration uses $K=15$ candidate labels and up to $M=5$ boundary
rules. Boundary retrieval changes the decision guidance supplied to the reader;
it does not change the candidate-label set itself.

\subsection{Lightweight Classification}
\label{app:reader}

The lightweight LLM receives the input text, the names and definitions of the
retrieved candidate labels, and the selected boundary rules. Retrieved
definitions provide semantic information about the candidate labels, while the
boundary rules provide input-relevant guidance for distinguishing potentially
confusable candidates. Retrieval scores and keyword lists are not included in
the final classification context.

The lightweight LLM performs a single online classification call and predicts
one label from the retrieved candidate set.

\section{Additional Dataset Details}
\label{app:datasets}

\subsection{Public Benchmarks}

We evaluate on three public benchmarks. WOS-46985 contains scientific abstracts
organized in a hierarchical taxonomy; the experiments use the fine-grained
Level-2 research-area names as classification labels, resulting in the
WOS-name-143 task. BANKING77 contains 77 fine-grained banking intents. CLINC150
contains 150 intent classes spanning multiple domains, and the experiments use
its in-scope examples.

The dataset splits used in the experiments are:

\begin{table}[h]
\centering
\caption{Dataset splits used in the experiments.}
\label{tab:app-datasets}
\small
\begin{tabular}{@{}lrl@{}}
\toprule
Dataset & Labels & Split \\
\midrule
WOS-name-143 & 143 & 28,191 train; 9,427 dev; 9,367 test \\
BANKING77 & 77 & 10,003 train; 3,080 test \\
CLINC150 & 150 & 15,000 train; 3,000 dev; 4,500 test \\
Book-38 & 38 & 540 train; 179 test \\
\bottomrule
\end{tabular}
\end{table}

SLKR knowledge construction uses only the label inventory and task description,
with no access to training, development, or test instances or classifier
feedback. In contrast, the few-shot and retrieved-ICL baselines use labeled
examples from the training split.

\subsection{Book-38}
\label{app:book38}

Book-38 is a real-world industrial dataset for large-scale book classification.
It contains 719 multilingual books categorized into 38 broad disciplines, with
540 training and 179 test instances. Compared with the public benchmarks,
Book-38 contains longer and more heterogeneous documents across a broader range
of subject areas and provides a complementary evaluation in a practical
data-curation setting.

\section{Additional Experimental Details}
\label{app:experiments}

\subsection{Models}

All retrieval-based methods use Qwen3-0.6B as the embedding model. We evaluate
two lightweight readers from different model families: Qwen3.5-4B and
Ministral-3-3B. The offline label knowledge used by SLKR is constructed with
GPT-5.6 Sol and frozen across the online experiments.

Knowledge construction uses only the label inventory and task description and
does not access training, development, or test instances or classifier
feedback. Few-shot and retrieved-ICL baselines use labeled examples from the
training split. Each experiment is repeated with three random seeds, and the
main paper reports mean performance across runs.

\subsection{Candidate Budget}

We use $K=15$ for SLKR across all four datasets. The first-stage retrieval
results reported in Table~3 of the main paper show that Recall@15 exceeds 95\%
on every dataset. We therefore fix $K=15$ for the subsequent classification and
boundary-retrieval experiments.

\subsection{Boundary-Retrieval Ablation}
\label{app:boundary-ablation}

The boundary-retrieval ablation fixes the candidate budget at $K=15$ and
compares four settings:

\begin{enumerate}
    \item \textbf{No Boundary ($M=0$):} boundary rules are removed.
    \item \textbf{All Boundaries:} all eligible rules connecting retrieved
    candidate labels are included.
    \item \textbf{Random $M=5$:} five eligible boundary rules are randomly
    selected.
    \item \textbf{Hybrid Top-$M=5$:} the five most input-relevant eligible
    boundary rules are selected using hybrid boundary retrieval.
\end{enumerate}

These settings separate the effect of explicit boundary knowledge from the
effect of input-conditioned boundary selection. As reported in Table~5, Hybrid
Top-$M=5$ uses approximately 0.35k boundary tokens per input, compared with
2.51k for All Boundaries. Random $M=5$ provides a rule-count-matched control.

\subsection{Evaluation Metrics}

Classification performance is evaluated using accuracy and macro-F1.
First-stage label retrieval is evaluated using Recall@$K$, measuring whether the
gold label is retained in the Top-$K$ candidates. Context efficiency is measured
by the average number of input tokens supplied to the lightweight reader.
Boundary-retrieval experiments additionally report the average number of tokens
contributed by selected boundary rules.

\section{Cost Accounting}
\label{app:cost}

SLKR separates one-time flagship knowledge-construction cost from recurring
online inference cost. Let $C_{\mathrm{offline}}$ denote the one-time
knowledge-construction cost and $c_{\mathrm{online}}$ the average recurring
online cost per classified input. For $T$ input instances, the total cost is
\[
C_{\mathrm{SLKR}}(T)
=
C_{\mathrm{offline}} + T c_{\mathrm{online}},
\]
and the corresponding amortized per-instance cost is
\[
\bar{C}_{\mathrm{SLKR}}(T)
=
\frac{C_{\mathrm{offline}} + T c_{\mathrm{online}}}{T}
=
\frac{C_{\mathrm{offline}}}{T}
+
c_{\mathrm{online}}.
\]

For a fixed label inventory, the contribution of the one-time construction cost
per input decreases as the number of classified inputs increases. Consequently,
\[
\lim_{T\rightarrow\infty}
\frac{C_{\mathrm{offline}}}{T}=0,
\qquad
\lim_{T\rightarrow\infty}
\bar{C}_{\mathrm{SLKR}}(T)=c_{\mathrm{online}}.
\]
Direct flagship classification, in contrast, requires a flagship-model inference
call for every input, so its inference cost grows linearly with $T$.

\end{document}